%% file: main.tex
\documentclass{article} 
\usepackage{iclr2027_conference,times}

\input{math_commands.tex}

\usepackage{hyperref}
\usepackage{url}
\usepackage{booktabs}
\usepackage{multirow}
\usepackage{subcaption}
\usepackage{graphicx} 
\usepackage{algorithm}
\usepackage{algpseudocode}
\usepackage{amssymb}
\usepackage{wrapfig}
\usepackage{capt-of}
\usepackage{tabularx}
\usepackage{float}

\title{Router Drift Control: Mitigating Off-Policy Degradation in Mixture-of-Experts Reinforcement Learning}

\newcommand{\rdcauthorblock}{%
{\centering\normalsize\bfseries
Liangyao Jin\textsuperscript{\normalfont 1,5,*}\quad
Zhixin Wang\textsuperscript{\normalfont 2,3,*}\quad
Liming Liu\textsuperscript{\normalfont 2,*}\quad
Mingjun Zhang\textsuperscript{\normalfont 5}\\
Zhengzhou Zhu\textsuperscript{\normalfont 1}\quad
Fangcheng Fu\textsuperscript{\normalfont 4}\quad
Jinlong Hou\textsuperscript{\normalfont 2}\quad
Yuan Cheng\textsuperscript{\normalfont 2}\quad
Xiaohe Hu\textsuperscript{\normalfont 5}\\
Ping Zhang\textsuperscript{\normalfont 5,$\dagger$}\quad
Siyuan Feng\textsuperscript{\normalfont 2,$\dagger$}\par}
{\centering\small
\textsuperscript{1}Peking University\quad
\textsuperscript{2}Shanghai Innovation Institute\quad
\textsuperscript{3}Zhejiang University\quad
\textsuperscript{4}Shanghai Jiaotong University\quad
\textsuperscript{5}Infrawaves Inc\par}
{\centering\small\rmfamily
zhangping@infrawaves.com\quad syfeng@sii.edu.cn\par}}

\author{\rdcauthorblock}

\makeatletter
\renewcommand{\@maketitle}{%
  \begingroup
  \raggedright
  {\LARGE\scshape\@title\par}
  \vspace{12pt}
  {\normalfont\@author\par}
  \vspace{16pt}
  \endgroup
}
\makeatother

\iclrfinalcopy 
\ClearShipoutPicture
\begin{document}

\enlargethispage{-24pt}
\AddToShipoutPictureFG*{%
  \AtTextLowerLeft{%
    \parbox[b]{\textwidth}{%
      \footnotesize
      \rule{0.3\textwidth}{0.4pt}\par
      \vspace{2pt}%
      \textsuperscript{*}Equal contribution; \textsuperscript{$\dagger$}Corresponding author.\par
    }%
  }%
}

\maketitle
\fancyhead{}

\input{paper/0-abstract}

\input{paper/1-introduction}

\input{paper/2-related}
\input{paper/3-method}
\input{paper/4-experiment}
\input{paper/5-conclusion}

\bibliography{conference}
\bibliographystyle{iclr2027_conference}

\input{paper/6-appendix}

\end{document}

%% file: math_commands.tex
\usepackage{amsmath,amsfonts,bm}

\def\eqref#1{equation~\ref{#1}}

\def\1{\bm{1}}

\DeclareMathAlphabet{\mathsfit}{\encodingdefault}{\sfdefault}{m}{sl}
\SetMathAlphabet{\mathsfit}{bold}{\encodingdefault}{\sfdefault}{bx}{n}



%% file: paper/0-abstract.tex
\begin{abstract}

Reinforcement learning (RL) can become off-policy due to training-inference mismatch, mini-batch updates, and asynchronous training.
The resulting mismatch is especially harmful in Mixture-of-Experts (MoE) models, where rollout--training discrepancies can change routing decisions and lead to router drift.
Rollout Routing Replay (R3) mitigates this mismatch by reusing expert selections from the rollout stage during training.
Our analysis shows that routing replay introduces gradient bias by losing contributions from experts selected by the training router but not the rollout router.
To address this, we propose \textbf{Router Drift Control (RDC)}, which uses routing disagreement between rollout and training as a signal of off-policy staleness and selectively rejects high-drift responses.
The sequence-level filtering avoids gradient bias introduced by replay while preserving native expert selection and the associated learning signals for retained responses.
Experiments show that RDC achieves larger gains as policy staleness increases and remains effective in asynchronous RL. In asynchronous training, RDC combined with Truncated Importance Sampling (TIS) improves the average peak validation accuracy across AIME 2024 and AIME 2025 by 5.5 points over TIS and 4.5 points over R3.

\end{abstract}

%% file: paper/1-introduction.tex
\section{Introduction}
\begin{figure*}[t]
    \centering
    \includegraphics[width=0.95\textwidth]{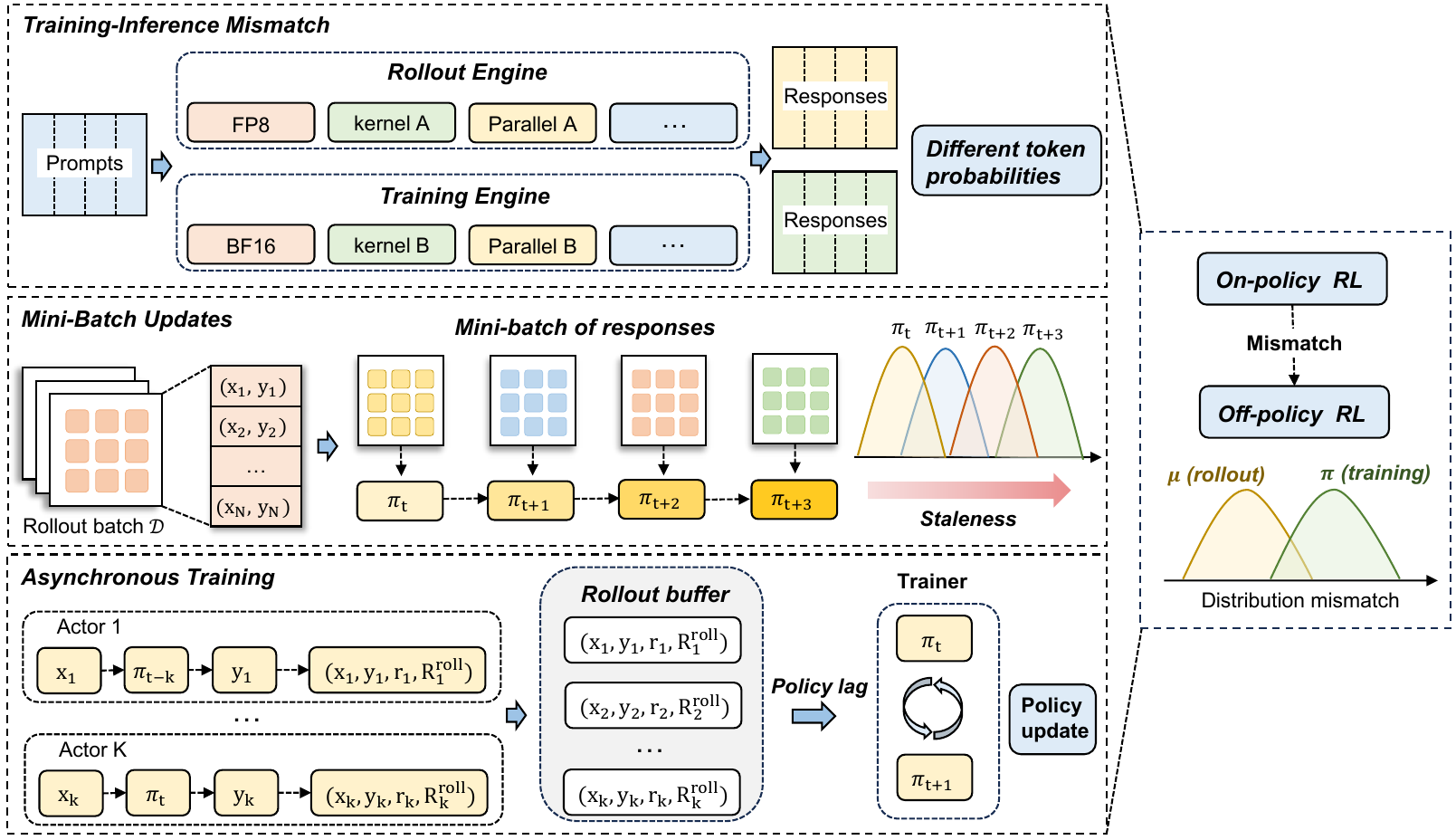}
    \caption{
    Three common sources of off-policy mismatch in LLM reinforcement learning. Training and inference discrepancy, sequential mini-batch updates, and asynchronous policy lag can cause the policy used for generation to diverge from the policy used for optimization.
    }
    \label{fig:off-policy-sources}
\end{figure*}

Reinforcement learning with verifiable rewards is widely used to improve the reasoning abilities of large language models~\citep{shao2024deepseekmath,deepseek2025r1,yu2025dapo}. 
In practical Reinforcement learning (RL) systems, rollout generation and policy optimization are often handled by separate inference and training engines~\citep{wang2026distflow}.
Consequently, the policy used to generate responses can differ from the current training policy, making nominally on-policy training effectively off-policy. 
As illustrated in Figure~\ref{fig:off-policy-sources}, off-policy mismatch can arise from three common sources:

\begin{itemize}
    \item \textbf{Training-Inference Mismatch.}
    Rollout and training engines may use different numerical precision, computational kernels, and parallel execution strategies~\citep{liu2025speed,vllm2025bitwise,he2025nondeterminism,zhang2025tbik}. As a result, even with identical model parameters and inputs, the rollout and training engines can produce different token probabilities.

    \item \textbf{Mini-Batch Updates.}
   Large rollout batches are often used to improve hardware utilization and are partitioned into multiple mini-batches for sequential policy updates~\citep{zheng2025gspo}.
   Because the policy is updated after each mini-batch, later mini-batches become increasingly stale relative to the current policy.

    \item \textbf{Asynchronous Training.}
    To improve system throughput by overlapping rollout generation with policy optimization, asynchronous RL decouples the two stages. As a result, training may use trajectories generated by earlier policy versions, introducing policy staleness
    ~\citep{espeholt2018impala,zhang2026relax,yao2025offpolicy}.
  \end{itemize}

\begin{figure*}[t]
    \centering
    \includegraphics[width=0.95\textwidth]{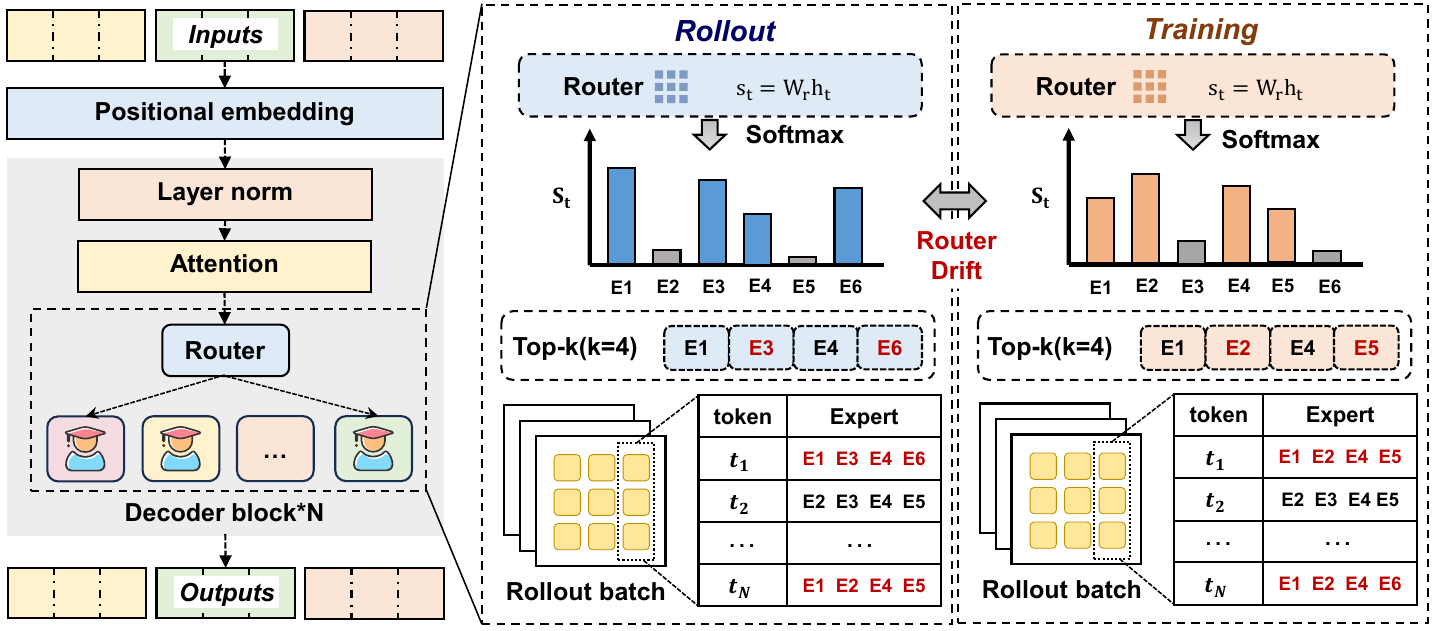}
    \caption{
    Router drift in MoE reinforcement learning. The same response can activate different experts during rollout and training, resulting in different computation paths and expert-level gradient attribution.
    }
    \label{fig:router-drift}
\end{figure*}

Mixture-of-Experts (MoE) model architectures scale model capacity by activating only a subset of experts for each token, with a router determining which experts participate in the computation ~\citep{shazeer2017sparsely,lepikhin2021gshard,fedus2022switch}. 
As a result, routing decisions directly determine the computation path and the experts that receive learning signals.
Under off-policy conditions, even small differences between the rollout and training policies can change routing decisions, causing the same token to be assigned to different experts during rollout and training ~\citep{ma2025r3,zhang2025rspo}. 
We refer to this rollout--training routing mismatch as \emph{router drift}.
As illustrated in Figure~\ref{fig:router-drift}, router drift changes the computation path of a response and the expert-level learning signals during training.

Existing methods mitigate off-policy effects from different perspectives ~\citep{zheng2025stabilizing,li2025trm,ma2025r3}.
Probability-based methods, such as truncated importance sampling (TIS), reweight token-level updates using truncated likelihood ratios between the training and rollout policies ~\citep{yao2025offpolicy,zheng2025stabilizing}. 
These ratios capture differences in token probabilities but do not directly reveal changes in expert assignments. 
Rollout Routing Replay (R3) addresses routing drift by reusing expert selections from the rollout stage during training~\citep{ma2025r3}.
While this enforces routing consistency between rollout and training, our analysis shows that replay alters gradient attribution relative to native routing, omitting contributions from experts that would otherwise be selected by the current training router.

We further observe that the degree of routing disagreement varies across responses.
Motivated by this insight, we propose Router Drift Control (RDC), which selectively filters high-drift responses while preserving the native routing of the remaining responses.
RDC quantifies routing disagreement between rollout and training. It uses this signal to exclude a small fraction of responses from each policy update. For retained responses, the training router remains free to select experts and receive the associated learning signals.
Our main contributions are summarized as follows.

\begin{itemize}
    \item 
    We analyze the role of routing in MoE reinforcement learning and show that routing decisions determine expert-level gradient attribution beyond what token probabilities can capture. We further show that routing replay alters this gradient attribution and can introduce bias relative to native routing.

    \item 
    We propose RDC, which quantifies rollout--training routing disagreement at the response level and uses it to identify high-drift responses. RDC selectively filters these responses while preserving the native training routes of retained responses, providing a routing-aware correction without route replay.

    \item 
    We evaluate RDC under both synchronous and fully asynchronous RL, showing that its gains become more pronounced as policy staleness increases. In asynchronous training, RDC combined with TIS improves the average peak validation accuracy on AIME 2024 and AIME 2025 by 5.5 points over TIS and 4.5 points over R3, while maintaining stable training dynamics.

\end{itemize}

%% file: paper/2-related.tex
\section{Related Work}

Modern LLM reinforcement learning systems often use separate engines
for rollout generation and policy optimization. Differences between these engines can cause rollout--training mismatch to produce different token probabilities even under identical model weights~\citep{yao2025offpolicy,liu2025speed,zheng2025stabilizing}. Recent work has shown that such mismatch can undermine the approximation of the sequence-level reward objective by token-level policy optimization and destabilize training~\citep{zheng2025stabilizing,zheng2025gspo}.
More broadly, rollout--training mismatch also arises from policy staleness. 
Multiple sequential updates and asynchronous rollout generation can create a gap between the training policy and the rollout policy ~\citep{schulman2017ppo,espeholt2018impala,zhang2026relax}. Existing work addresses these sources of mismatch at the system, probability, and routing levels.

\subsection{System-Level Corrections}

System-level methods aim to reduce training--inference mismatch
by improving  numerical consistency between the rollout and training engines. \citet{he2025nondeterminism} study inference nondeterminism caused by non-associative floating-point reductions and batch-dependent execution. \citet{zhang2025tbik} develop tensor-parallel-invariant kernels to maintain bitwise consistency across different parallel configurations. Related work also aligns training and inference implementations to support bitwise-consistent on-policy RL~\citep{vllm2025bitwise}. 

These methods mitigate mismatch arising from numerical differences between rollout and training. 
However, numerical consistency alone cannot address policy staleness caused by sequential mini-batch updates or asynchronous training. 
Algorithm-level corrections are therefore still needed.

\subsection{Probability-Space Corrections}

Algorithm-level methods address rollout--training mismatch by modifying how rollout data contributes to policy updates. 
A common approach is to control token-level importance ratios. 
TIS truncates large ratios~\citep{yao2025offpolicy,zheng2025stabilizing}, while CISPO  stops gradients through the importance weights ~\citep{minimax2025m1}. 
Other methods modify how these ratios are aggregated:  
GSPO uses sequence-level ratios~\citep{zheng2025gspo}, GMPO adopts geometric rather than arithmetic averaging~\citep{zhao2026gmpo}, and CTPO accounts for cumulative ratios across tokens~\citep{zhang2026ctpo}. 
Some methods instead apply probability-based corrections selectively. 
SIS adjusts the contribution of off-policy tokens~\citep{li2026sis}, while TRM identifies trust-region violations using the maximum token-level KL divergence and masks the corresponding sequences~\citep{li2025trm}.

Although these methods use different objectives and correction rules, they primarily characterize off-policy mismatch through probability-space signals, such as token probabilities and likelihood ratios. These signals can capture distributional differences between rollout and training, but they do not directly reflect the computation paths taken by an MoE model. This limitation becomes important when the same response is routed to different experts during rollout and training.

\subsection{Routing Corrections for MoE Models}

Sparse MoE models increase capacity by activating only a subset of experts for each token ~\citep{shazeer2017sparsely,lepikhin2021gshard,fedus2022switch}. 
Their discrete routing decisions can be sensitive to small computational changes~\citep{tranhuu2026geometry}, making rollout--training routing mismatch an important issue in MoE reinforcement learning. 
\citet{ma2025r3} study routing disagreement between rollout and training and introduces two forms of routing replay. Routing Replay (R2) records the routes selected by the training policy during probability computation and reuses them during the policy update. R3 further reuses rollout-time routes in the training forward pass~\citep{ma2025r3}. PR$^2$ predicts updated routing decisions to reduce the staleness caused by replaying fixed rollout routes~\citep{dong2026pr2}. RSPO uses router-aware importance weights to account for changes in routing scores~\citep{zhang2025rspo}.

These methods address routing disagreement either by modifying the training routes or by reweighting rollout data. In doing so, they alter either the computation path or the contribution of each response to the policy update.
RDC takes a different approach: it uses response-level router drift to determine which responses participate in optimization, while preserving the native training routes of retained responses. Combined with TIS, RDC addresses routing disagreement and token-level probability mismatch through complementary mechanisms.

%% file: paper/3-method.tex
\section{Method}
\label{sec:method}

\subsection{Setup and Notation}
\label{sec:setup}

We consider reinforcement learning with separate rollout and training engines. Let $\mu(\cdot\mid x)$ denote the behavior policy realized by the rollout engine, and let $\pi_\theta(\cdot\mid x)$ denote the current training policy.

For a sampled response $y=(y_1,\ldots,y_T)$, let $\mathcal{T}=\{1,\ldots,T\}$ denote its token positions. We consider an MoE model with $L$ MoE layers and $E$ experts per layer. Each token is routed to $K$ experts. For token $t$ at layer $l$, we denote the expert sets selected during rollout and training by
\begin{equation}
R_l^\mu(t), R_l^\pi(t) \subseteq \{1,\ldots,E\},
\qquad
|R_l^\mu(t)|=|R_l^\pi(t)|=K.
\end{equation}
We use $R^\mu(y)$ and $R^\pi(y)$ to denote the routing assignments over
the full response.

The training route is obtained by evaluating the sampled response under
teacher forcing.
Let $c_t=(x,y_{<t})$ denote the context before token $y_t$.
The rollout and training engines assign probabilities
$\mu(y_t\mid c_t)$ and $\pi_\theta(y_t\mid c_t)$, respectively.

\subsection{Router Drift Control}
\label{sec:rdc}

RDC uses rollout--training routing disagreement to decide which responses should contribute to a policy update. RDC does not force the training engine to replay rollout routes. Instead, it filters responses with large router drift and preserves the current training policy's routing decisions for retained responses.

\begin{figure*}[t]
    \centering
    \includegraphics[width=0.95\textwidth]{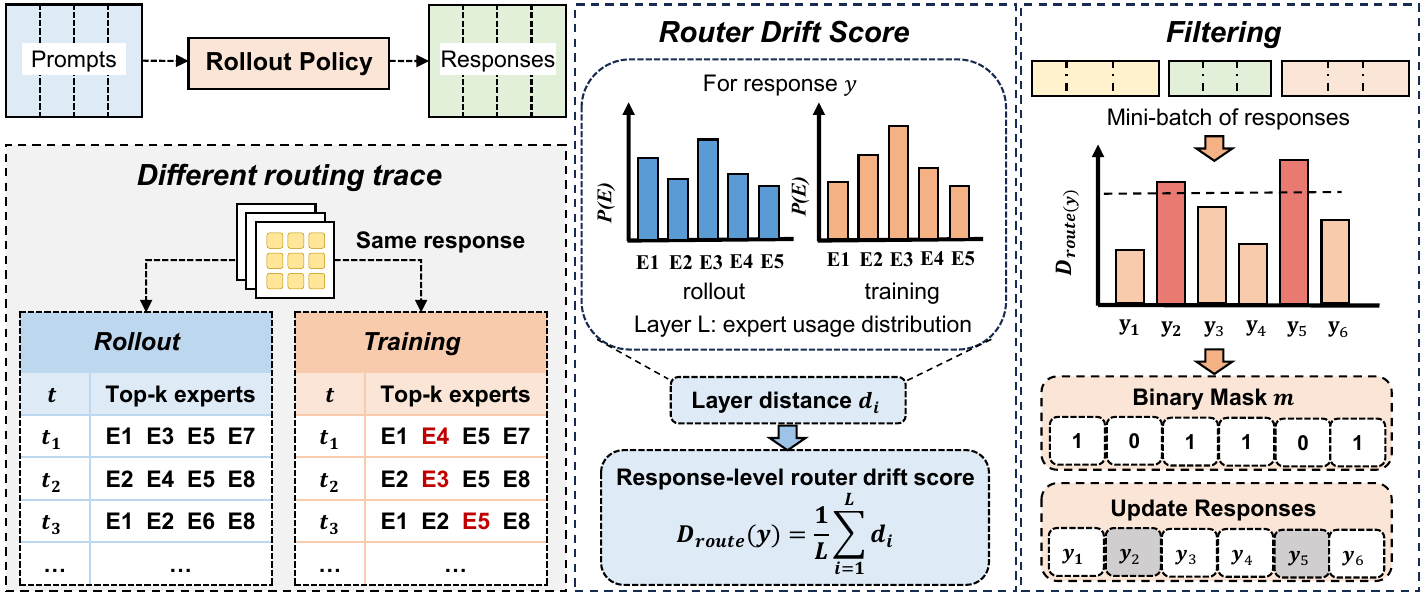}
    \caption{
    Overview of Router Drift Control (RDC). RDC aggregates layer-wise differences in expert usage between rollout and training into a response-level drift score. It filters high-drift responses before policy updates. Retained responses follow their native training routes.
    }
    \label{fig:rdc-overview}
\end{figure*}

Figure~\ref{fig:rdc-overview} shows the overall procedure.
For each response, RDC compares expert-usage distributions between rollout and training across MoE layers. It aggregates the layer-wise differences into a response-level router drift score. Based on this score, RDC filters a small fraction of responses with the largest drift before policy optimization.

\paragraph{Response-level router drift.}

For engine $a\in\{\mu,\pi\}$, we define the expert-usage distribution
at layer $l$ over response $y$ as
\begin{equation}
u_l^{a}(e\mid y)
=
\frac{1}{TK}
\sum_{t=1}^{T}
\mathbb{I}\left[e\in R_l^{a}(t)\right],
\qquad
e\in\{1,\ldots,E\}.
\label{eq:expert-usage}
\end{equation}

Here, $u_l^{a}(e\mid y)$ denotes the fraction of expert selections assigned to expert $e$, normalized over all $TK$ selections in the response. Thus, $u_l^{a}(\cdot\mid y)$ forms a distribution over experts.

We measure the routing difference at each layer between the two expert-usage distributions using total variation distance:
\begin{equation}
d_l(y)
=
\frac{1}{2}
\left\|
u_l^\mu(\cdot\mid y)
-
u_l^\pi(\cdot\mid y)
\right\|_1.
\label{eq:layer-router-drift}
\end{equation}

We then average the layer-wise distances to obtain the response-level router drift score:
\begin{equation}
D_{\mathrm{route}}(y)
=
\frac{1}{L}
\sum_{l=1}^{L} d_l(y).
\label{eq:route-drift-score}
\end{equation}

The score lies in $[0,1]$, with larger values indicating greater mismatch in aggregate expert usage between rollout and training. It characterizes the overall shift in expert utilization across the response.

\paragraph{Selective response filtering.}

Let $q\in[0,1)$ denote the filtering ratio.
RDC ranks responses by $D_{\mathrm{route}}(y)$ and filters those with the
highest scores.

In practice, we rank responses by their router drift scores within groups of similar response lengths, rather than comparing scores across substantially different lengths.

For a rollout batch $\mathcal{B}$, let $\mathcal{F}_q(\mathcal{B})$
denote the responses selected for filtering. The response-level mask is
\begin{equation}
m_{\mathcal{B}}(y)
=
\mathbb{I}\left[
y\notin\mathcal{F}_q(\mathcal{B})
\right].
\label{eq:response-mask}
\end{equation}

\setlength{\intextsep}{2pt}
\begin{wrapfloat}{algorithm}[14]{r}{0.48\textwidth}
\footnotesize
\caption{Router Drift Control}
\label{alg:rdc}
\begin{algorithmic}[1]
\Require Batch $\mathcal{B}$, rollout routes $R^\mu$,
training policy $\pi_\theta$, ratio $q$
\Ensure Updated policy $\pi_\theta$

\ForAll{$y\in\mathcal{B}$}
    \State Obtain training routes $R^\pi(y)$
    \State Compute $D_{\mathrm{route}}(y)$
    using Eqs.~\ref{eq:expert-usage}--\ref{eq:route-drift-score}
\EndFor

\State Partition $\mathcal{B}$ into length buckets
\ForAll{buckets $\mathcal{B}_b$}
    \State Select the top
    $\lceil q|\mathcal{B}_b|\rceil$ responses
\EndFor

\State Construct $m_{\mathcal{B}}$
using Eq.~\ref{eq:response-mask}
\State Compute $\mathcal{L}_{\mathrm{RDC}}(\theta)$
using Eq.~\ref{eq:rdc-update}
\State Update $\theta$ using
$\nabla_\theta\mathcal{L}_{\mathrm{RDC}}(\theta)$

\end{algorithmic}
\end{wrapfloat}

Let $\ell_\theta(y;R)$ denote the sum of the token-level
policy losses for response $y$ under routing assignment $R$,
and let $N_{\mathcal B}$ denote the total number of valid tokens
in the retained responses. The RDC objective is
\begin{equation}
\mathcal{L}_{\mathrm{RDC}}(\theta)
=
\frac{1}{N_{\mathcal B}}
\sum_{y\in\mathcal B}
m_{\mathcal B}(y)\,
\ell_\theta\!\left(y;R^\pi(y)\right).
\label{eq:rdc-update}
\end{equation}

The mask excludes filtered responses from the policy update while retained responses follow the routing decisions of the current training policy. RDC can be combined with token-level probability corrections such as TIS. 
Algorithm~\ref{alg:rdc} describes a single optimizer update and is applied separately to each update mini-batch. Before every update, we recompute the routing scores and filtering mask using the current training policy, and rank responses within the length buckets of the current mini-batch.

\subsection{Routing Determines Gradient Support}
\label{sec:routing-gradient}

RDC uses routing information because expert selection determines which parameters can receive the local training signal.

At layer $l$ and token $t$, an MoE layer computes
\begin{equation}
h_l^{\mathrm{out}}(t)
=
\sum_{e\in R_l(t)}
\operatorname{gate}_{l,e}(t)
E_{l,e}\!\left(h_l^{\mathrm{in}}(t)\right),
\label{eq:moe-layer}
\end{equation}
where $E_{l,e}$ denotes expert $e$ and $\operatorname{gate}_{l,e}(t)$ denotes its routing weight~\citep{shazeer2017sparsely,fedus2022switch}.

Let $\theta_{l,e}$ denote the parameters of expert $e$.
If the router does not select this expert, the token provides
no direct gradient to its parameters through this MoE layer:
\begin{equation}
e\notin R_l(t)
\quad\Longrightarrow\quad
\frac{\partial h_l^{\mathrm{out}}(t)}
     {\partial \theta_{l,e}}
=
0.
\label{eq:routing-gradient-support}
\end{equation}
Thus, routing determines which expert parameter blocks can receive direct gradients through the current token's MoE computation.

This matters when rollout and training use different routes. The rollout engine processes response $y$ through $R^\mu(y)$. The training engine computes gradients through $R^\pi(y)$. If these routes differ, the expert support used during training differs from the one used during rollout. Some experts activated during rollout may receive no direct gradient in training. Other experts may receive gradients even though they were not used during
rollout. 

Router drift can therefore change how learning signals are distributed across experts, even when its effect on token probabilities is small.

\subsection{Token Probabilities Do Not Identify Routing}
\label{sec:probability-limit}

Token probabilities do not uniquely determine the underlying routing decisions. Different expert paths can produce the same output distribution while assigning gradients to different expert parameter blocks.
Consider a top-$K$ MoE layer. Let $S$ contain $K-1$ shared experts, and let $e_1$ and $e_2$ be two distinct experts outside $S$.
Consider two routing assignments
\begin{equation}
R_1=S\cup\{e_1\},
\qquad
R_2=S\cup\{e_2\}.
\label{eq:route-construction}
\end{equation}

Suppose the shared experts have the same routing weights under both assignments, and the two replacing experts have equal routing weights. Further suppose that their outputs at the current input are identical:
\begin{equation}
E_{l,e_1}\!\left(h_l^{\mathrm{in}}(t)\right)
=
E_{l,e_2}\!\left(h_l^{\mathrm{in}}(t)\right).
\label{eq:equal-expert-output}
\end{equation}
Under these conditions, the two routing assignments produce the same MoE output. If the remaining computation is identical, they also produce the same logits and probability distribution. Consequently, likelihood ratios and token-level KL divergence computed from these probabilities cannot identify which routing assignment was used.

The direct expert-gradient support can nevertheless differ.
Under $R_1$, expert $e_1$ participates in the MoE computation,
whereas expert $e_2$ does not. Under $R_2$, their roles are reversed:
\begin{equation}
\frac{\partial h_l^{\mathrm{out}}(t;R_1)}
     {\partial \theta_{l,e_2}}
=
0,
\qquad
\frac{\partial h_l^{\mathrm{out}}(t;R_2)}
     {\partial \theta_{l,e_1}}
=
0.
\label{eq:route-gradient-difference-example}
\end{equation}
Thus, even when the two assignments produce identical token
probabilities, they can direct learning signals to different
expert parameter blocks. Routing information therefore reveals
differences in the training computation that cannot be uniquely
recovered from token probabilities alone.

\subsection{Selective Control versus Route Replay}
\label{sec:selective-correction}

Probability-based corrections, such as TIS, reweight rollout data while preserving the native training route. R3 instead replaces this route with the route recorded during rollout~\citep{ma2025r3}.

To examine the effect of route substitution, let
\begin{equation}
g_\theta(y;R)
=
\nabla_\theta \ell_\theta(y;R)
\label{eq:gradient-contribution}
\end{equation}
denote the gradient contribution of response $y$ under routing assignment $R$, where $\ell_\theta(y;R)$ is defined in Section~\ref{sec:rdc}. For the same response, replacing the training route with the rollout route changes the gradient contribution by
\begin{equation}
\Delta_{\mathrm{route}}(y)
=
g_\theta\!\left(y;R^\mu(y)\right)
-
g_\theta\!\left(y;R^\pi(y)\right).
\label{eq:route-gradient-shift}
\end{equation}

The substitution can alter expert-level gradient support and introduce bias relative to the native-routing objective. Appendix~\ref{R3_analysis} provides a detailed analysis.

RDC retains the native training route and uses routing disagreement to determine which responses contribute to the update. We combine RDC with TIS: TIS adjusts token-level contributions using truncated importance weights, while RDC filters responses based on router drift.

%% file: paper/4-experiment.tex
\section{Experiments}
We evaluate RDC on mathematical and logical reasoning tasks under both synchronous and asynchronous RL settings. In synchronous training, we vary the update batch size to study the effects of sequential policy updates on rollout data. 
We further evaluate RDC under fully asynchronous training with policy lag between rollout and optimization.
We compare GRPO, TIS, R3, and RDC combined with TIS, using validation accuracy as the primary metric.
For simplicity, we use RDC to denote RDC combined with TIS in the experiments below.
We also track training reward, gradient norm, rollout--training KL divergence, and response length to analyze training dynamics.

\subsection{Experimental Setup}
\textbf{Model and Datasets:}  
We train Qwen3-30B-A3B~\citep{yang2025qwen3} on DAPO-Math-17K~\citep{yu2025dapo}. Evaluation uses AIME 2024, AIME 2025, MATH-500~\citep{lightman2023lets} and ZebraLogic~\citep{lin2025zebralogic}. All methods follow the same evaluation protocol on each benchmark. We evaluate every five global steps and report the best validation accuracy reached during training. 

\textbf{Methods:}
All methods use GRPO~\citep{shao2024deepseekmath} as the baseline, with DAPO Clip-Higher objective~\citep{yu2025dapo}  ($\epsilon_{\mathrm{low}}=0.2$,
$\epsilon_{\mathrm{high}}=0.27$).
We compare four settings: GRPO, R3, TIS, and RDC+TIS. RDC ranks responses by $D_{\mathrm{route}}(y)$ within response-length groups and filters the highest-drift fraction before optimization. Unless otherwise specified, we set the filtering ratio to $q=0.08$. We combine RDC with TIS to address routing disagreement and token-level probability mismatch: RDC filters responses with large routing disagreement, while TIS corrects token-level probability mismatch on the retained responses.
Within each training regime, all methods use the same rollout budget and
training configuration.

\textbf{Multiple Mini-Steps:}
In synchronous training, each rollout batch contains 2,048 responses,
generated from 256 prompts with $n=8$ responses per prompt. We keep the rollout batch size fixed and vary the number of responses used in each policy update. Specifically, we consider $(\mathcal{B}_{\mathrm{rollout}}, \mathcal{B}_{\mathrm{update}}) \in \{(2048,2048), (2048,512), (2048,128)\}$. A smaller update batch leads to more successive updates on the same rollout
batch, increasing the staleness of the data relative to the current policy.

\textbf{Asynchronous Training:}
We further evaluate RDC under fully asynchronous training. Rollout generation and policy optimization run independently, and the Trainer consumes generated samples from a rollout queue. Each prompt generates $n=8$ responses, and each local update uses 32 prompt groups, corresponding to 256 response trajectories. The rollout and training processes therefore continue at different rates, allowing the trainer to consume data generated by earlier policy versions. 
For each training batch, the Trainer recomputes routing under the current policy, measures router drift, and filters high-drift responses before the policy update.
After every four local updates, the updated Trainer parameters are synchronized to the rollout workers.

\textbf{Other Settings:}
We use Megatron for BF16 policy optimization and vLLM for FP8 rollout
generation~\citep{shoeybi2019megatron,kwon2023vllm}.
We set the maximum response length to 20,480 tokens during training and
30,720 tokens during evaluation.
Additional optimization, asynchronous scheduling, batching, and
implementation details are provided in Appendix~\ref{implementation-details}.

\begin{table*}[t]
  \centering
  \small
  \caption{
  Best validation accuracy under different update batch sizes in synchronous training. }
  \label{tab:sync-results}
  \begin{tabular}{llcccc}
  \toprule
  $(\mathcal{B}_{\mathrm{rollout}},\mathcal{B}_{\mathrm{update}})$
  & Method & ZebraLogic & AIME24 & AIME25 & MATH-500 \\
  \midrule
  \multirow{4}{*}{$(2048,2048)$}
  & GRPO & 61.2 & 77.7 & 65.2 & 96.6 \\
  & GRPO+R3 & 63.0 & 74.9 & 65.4 & 97.6 \\
  & GRPO+TIS & 66.1 & \textbf{80.0} & 71.3 & \textbf{98.2} \\
  & GRPO+TIS+RDC & \textbf{73.9} & 78.5 & \textbf{72.4} & 97.8 \\
  \midrule
  \multirow{4}{*}{$(2048,512)$}
  & GRPO & 46.3 & 62.2 & 50.8 & 96.0 \\
  & GRPO+R3 & 55.2 & 69.9 & 57.9 & 96.6 \\
  & GRPO+TIS & 61.3 & 70.6 & 60.3 & 97.0 \\
  & GRPO+TIS+RDC & \textbf{68.1} & \textbf{71.5} & \textbf{62.5} & \textbf{97.6} \\
  \midrule
  \multirow{4}{*}{$(2048,128)$}
  & GRPO & 61.1 & 53.4 & 40.0 & 92.2 \\
  & GRPO+R3 & 53.1 & 61.0 & 48.7 & 95.0 \\
  & GRPO+TIS & 78.2 & 70.6 & 62.9 & 97.6 \\
  & GRPO+TIS+RDC & \textbf{86.1} & \textbf{77.0} & \textbf{67.2} & \textbf{97.8} \\
  \bottomrule
  \end{tabular}
  \end{table*}

\subsection{Experimental Results}
\paragraph{Synchronous Training.}
We first compare the four methods under synchronous training with different update batch sizes. 
Table~\ref{tab:sync-results} reports the best validation accuracy observed during training.
RDC performs competitively under low policy staleness and delivers larger gains as the update batch becomes smaller. Under $(2048,512)$, RDC achieves the highest accuracy on all four benchmarks. The advantage becomes more pronounced under $(2048,128)$.
Compared with TIS alone, RDC improves ZebraLogic, AIME24, and AIME25 by
7.9, 6.4, and 4.3 percentage points, respectively. These results suggest that routing-aware response selection becomes more useful as rollout data grows increasingly stale.

  \begin{table}[t]
  \centering
  \small
  \caption{Best validation accuracy under asynchronous RL.}
  \label{tab:async-results}
  \begin{tabular}{lcccc}
  \toprule
  Method & ZebraLogic & AIME24 & AIME25 & MATH-500 \\
  \midrule
  GRPO & 46.9 & 63.4 & 50.7 & 96.2 \\
  GRPO+R3 & \textbf{53.9} & 68.7 & 53.1 & 96.2 \\
  GRPO+TIS & 53.7 & 65.0 & 54.8 & 96.4 \\
  GRPO+TIS+RDC & 53.0 & \textbf{71.2} & \textbf{59.6} & \textbf{97.2} \\
  \bottomrule
  \end{tabular}
  \end{table}

\begin{figure*}[t]
      \resizebox{\textwidth}{!}{%
    \includegraphics[height=3cm]{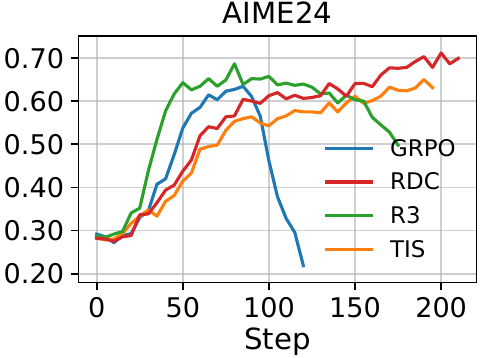}\hspace{4pt}
    \includegraphics[height=3cm]{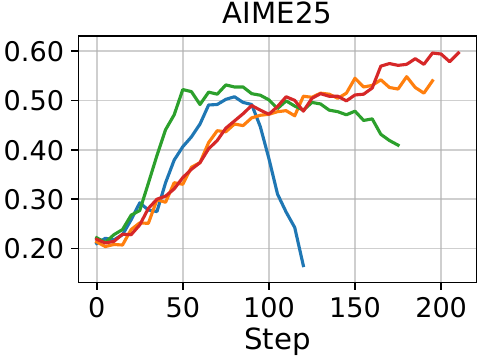}\hspace{4pt}
    \includegraphics[height=3cm]{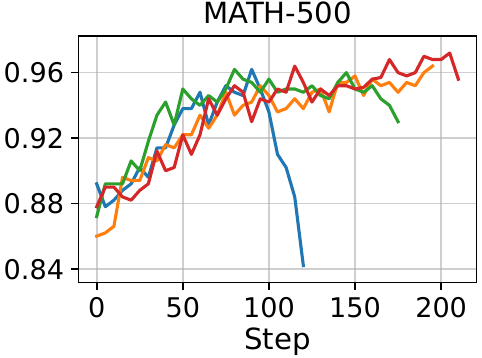}\hspace{4pt}
    \includegraphics[height=3cm]{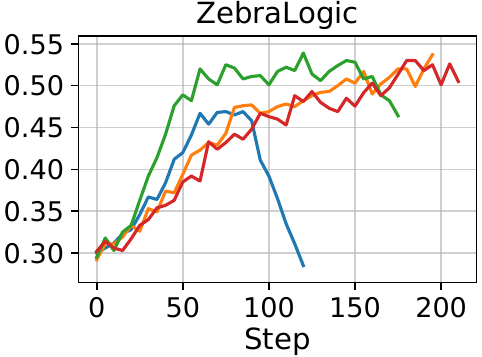}%
  }
      \caption{Validation accuracy during asynchronous reinforcement learning
      on AIME24, AIME25, MATH-500, and ZebraLogic.}
      \label{fig:async-validation}
  \end{figure*}

\paragraph{Asynchronous Training.}
We further evaluate RDC under fully asynchronous training.
Table~\ref{tab:async-results} reports the best validation accuracy, while
Figure~\ref{fig:async-validation} shows the full validation trajectories.
RDC achieves the highest accuracy on AIME24, AIME25, and MATH-500, while R3 performs slightly better on ZebraLogic. On AIME24 and AIME25, RDC improves the average peak score by 5.5 points over TIS and 4.5 points over R3.  R3 avoids reward collapse but exhibits late-stage degradation in validation accuracy, whereas RDC maintains strong late-stage performance, particularly on the two AIME benchmarks. The uncorrected GRPO baseline collapses after roughly 100 training steps, while RDC continues to improve and remains stable across the evaluation benchmarks.



\begin{figure*}[t]
      \resizebox{\textwidth}{!}{%
    \includegraphics[height=3cm]{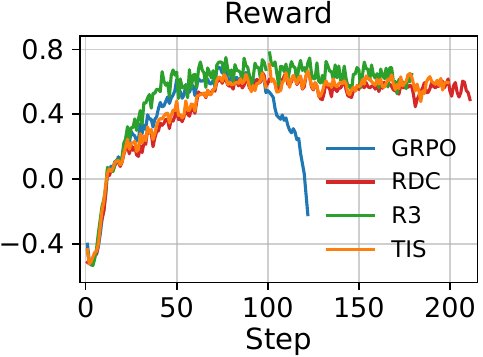}\hspace{4pt}
    \includegraphics[height=3cm]{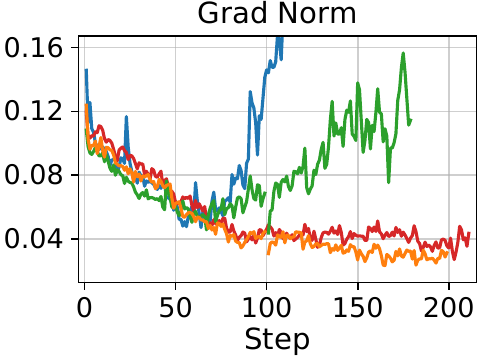}\hspace{4pt}
    \includegraphics[height=3cm]{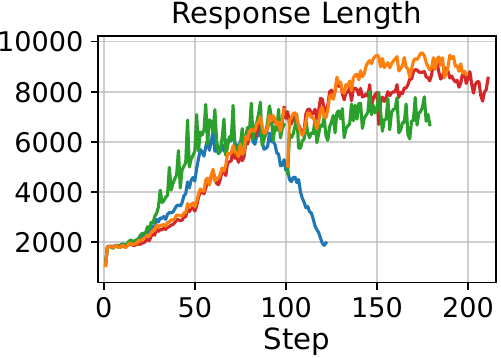}\hspace{4pt}
    \includegraphics[height=3cm]{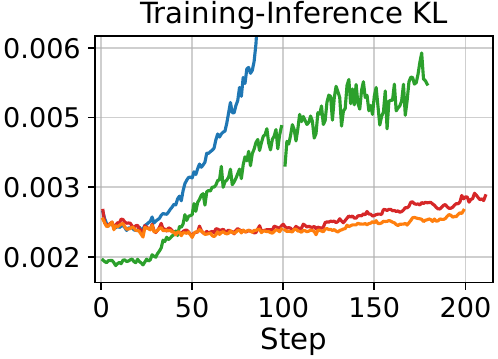}%
  }
      \caption{Training dynamics under asynchronous reinforcement learning, showing training reward, gradient norm, response length, and training--inference KL for GRPO, TIS, R3, and RDC+TIS.
    }
    \label{fig:async-dynamics}
  \end{figure*}

\paragraph{Training Dynamics.}

Figure~\ref{fig:async-dynamics} provides a closer view of the training
dynamics under asynchronous RL. The uncorrected GRPO baseline exhibits clear late-stage degradation. Its reward rises during early training but drops sharply after around 100 steps and eventually becomes negative.
The response length decreases at the same time, from roughly 6,000 tokens
to around 2,000 tokens. In contrast, TIS, R3, and RDC maintain positive rewards and avoid this response-length collapse. This degradation coincides with rapid growth in both training--inference KL and gradient norm. 

For GRPO, the training--inference KL increases rapidly and eventually exceeds the displayed range, while the gradient norm rises sharply during the same stage. TIS and RDC keep both quantities low and stable throughout training, whereas R3 shows a more pronounced increase in both KL and gradient norm during later stages.

Overall, RDC improves validation accuracy over TIS under smaller update batches and asynchronous training, while maintaining stable training dynamics.

%% file: paper/5-conclusion.tex
\section{conclusion}

Off-policy mismatch can degrade performance and destabilize
reinforcement learning in MoE models. In this work, we study router drift as a distinct source of off-policy mismatch in MoE reinforcement learning. We show that changes in routing alter the expert-level gradient support and that replaying rollout routes changes the training gradient relative to the native routing of the current policy. Based on this observation, we propose Router Drift Control (RDC), which filters high-drift responses while preserving the current training policy's routing decisions. Combined with TIS, RDC improves validation performance and training stability under mini-batch updates and fully asynchronous training. These results demonstrate the value of routing-aware filtering as an alternative to rollout routing replay.

%% file: paper/6-appendix.tex
\newpage
\onecolumn
\appendix

\section{Experimental Details}
\label{implementation-details}

Table~\ref{tab:implementation-details} summarizes additional experimental hyperparameters, followed by implementation details for RDC and the two training regimes.

\begin{table}[t]
\centering
\small
\caption{Additional experimental hyperparameters. TP, PP, EP, and ETP denote tensor, pipeline, expert, and expert tensor parallelism, respectively.}
\label{tab:implementation-details}
\begin{tabular}{@{}ll@{}}
\toprule
Parameter & Value \\
\midrule
Learning rate (synchronous / asynchronous) & $3\times10^{-6}$ / $1\times10^{-6}$ \\
Weight decay / gradient clipping threshold & $0.1$ / $1.0$ \\
PPO epochs & $1$ \\
Loss aggregation & \texttt{token-mean} \\
Sampling temperature (training / validation) & $1.0$ / $0.6$ \\
Top-$p$ (training / validation) & $1.0$ / $0.95$ \\
Top-$k$ (training / validation) & Disabled / $20$ \\
TIS threshold $C$ & $2.0$ \\
Response-length bucket boundaries & $2048,\ 4096,\ 8192,\ 12288$ \\
Megatron TP / PP / EP / ETP & $4$ / $1$ / $4$ / $2$ \\
\bottomrule
\end{tabular}
\end{table}

We initialize all runs from the post-trained Qwen3-30B-A3B
checkpoint and disable thinking mode during both training
rollouts and evaluation. For AIME 2024 and AIME 2025, we sample $32$ responses per problem and compute accuracy by averaging binary correctness over the sampled responses and problems. During training, a response receives a reward of $+1$ if its answer is correct and $-1$ otherwise. During evaluation, correctness is scored as $1$ for a correct answer and $0$ otherwise.

For asynchronous training, rollout generation and policy optimization use separate resource pools. We set \texttt{require\_batches} to $1$ and \texttt{staleness\_threshold} to $0.5$. For each Trainer batch, RDC computes the response-level expert-usage difference once, using the BF16 routing assignments recorded when computing \texttt{old\_log\_prob}. It then filters responses before the policy update.

\section{Gradient Bias under Rollout Routing Replay}
\label{R3_analysis}

We analyze the gradient bias introduced by rollout routing replay (R3)~\citep{ma2025r3}. We first derive the local gradients of an MoE layer and then compare the gradients and optimization objectives under replayed and native routing.

\subsection{Setup}

We consider a sparsely gated MoE layer~\citep{shazeer2017sparsely}. For one token at this layer, we omit the token and layer indices for clarity. Let $h$ be the layer input, and let
\begin{equation}
z = W_r h + b_r,
\qquad
S^{\mathrm{cur}} = \operatorname{TopK}(z).
\end{equation}
Native routing selects the current expert set $S^{\mathrm{cur}}$. In contrast, R3 uses the expert set $S^{\mathrm{roll}}$ recorded during rollout.

We consider routing replay that fixes expert identities but recomputes gate weights from the current router logits. For a selected set $S$, let
\begin{equation}
g_e^S=
\frac{\exp(z_e)}{\sum_{i\in S}\exp(z_i)},
\qquad e\in S.
\label{eq:r3-normalized-gate}
\end{equation}
The MoE output is
\begin{equation}
o^S=
\sum_{e\in S}g_e^S E_e(h;\psi_e),
\label{eq:r3-moe-output}
\end{equation}
where $\psi_e$ denotes the parameters of expert $e$.

Our analysis considers local gradients through this MoE output, holding the layer input fixed. Expert parameters are distinct, and expert outputs do not depend directly on router logits. We exclude auxiliary router losses and assume no token dropping. As in standard top-$k$ routing, the selected expert set is treated as fixed during backpropagation. The derivatives below therefore apply away from routing selection boundaries.

\subsection{Gradients Under Fixed Routing}

Let $\mathcal{L}^S$ be the downstream loss evaluated
using expert set $S$, and define
\begin{equation}
\delta^S=
\nabla_{o^S}\mathcal{L}^S,
\qquad
a_e^S=(\delta^S)^\top E_e(h;\psi_e).
\end{equation}
Both $o^S$ and $\delta^S$ may depend on the selected expert set.

For $e,j\in S$, differentiating the normalized gate weights gives
\begin{equation}
\frac{\partial g_e^S}{\partial z_j}
=
g_e^S\bigl(\mathbf{1}[e=j]-g_j^S\bigr).
\end{equation}
Consequently,
\begin{equation}
\frac{\partial o^S}{\partial z_j}
=
g_j^S\bigl(E_j(h;\psi_j)-o^S\bigr),
\qquad j\in S.
\end{equation}
For $j\notin S$, the output has no direct dependence
on $z_j$ when $S$ is held fixed. Thus, the local
router-logit gradient is
\begin{equation}
r_j^S
\equiv
\frac{\partial\mathcal{L}^S}{\partial z_j}
=
\begin{cases}
g_j^S
\left(
a_j^S-\displaystyle\sum_{e\in S}g_e^S a_e^S
\right), & j\in S,\\
0, & j\notin S.
\end{cases}
\label{eq:r3-logit-gradient}
\end{equation}

For the linear router, the corresponding local
parameter gradients are
\begin{equation}
\left.\nabla_{W_r}\mathcal{L}^S\right|_{\mathrm{local}}
=r^S h^\top,
\qquad
\left.\nabla_{b_r}\mathcal{L}^S\right|_{\mathrm{local}}
=r^S.
\end{equation}

Let
$J_j(h)=\partial E_j(h;\psi_j)/\partial\psi_j$
denote the expert Jacobian. The direct expert
gradient is
\begin{equation}
\left.
\nabla_{\psi_j}\mathcal{L}^S
\right|_{\mathrm{local}}
=
\begin{cases}
g_j^S J_j(h)^\top\delta^S,
& j\in S,\\
0,
& j\notin S.
\end{cases}
\label{eq:r3-direct-expert-gradient}
\end{equation}
These are local gradient contributions for one token at one layer. An unselected expert may still receive gradients from other tokens.

\subsection{Changes in Local Gradient Support}

\paragraph{Proposition 1.}
Fix the layer input and current model parameters. Suppose
$j\in S^{\mathrm{cur}}\setminus S^{\mathrm{roll}}$.
Routing replay gives zero local gradients with
respect to $z_j$ and $\psi_j$. Under native routing,
the direct expert gradient is nonzero if
\begin{equation}
J_j(h)^\top\delta^{\mathrm{cur}}\ne 0,
\end{equation}
and the local router-logit gradient is nonzero if
\begin{equation}
(\delta^{\mathrm{cur}})^\top
\left(E_j(h;\psi_j)-o^{\mathrm{cur}}\right)
\ne 0,
\end{equation}
where the superscript $\mathrm{cur}$ denotes evaluation under $S^{\mathrm{cur}}$.

\paragraph{Proof.}
Since $j\notin S^{\mathrm{roll}}$, the replay output has no direct dependence on $z_j$ or $\psi_j$. Both local gradients are therefore
zero by Eqs.~\ref{eq:r3-logit-gradient} and~\ref{eq:r3-direct-expert-gradient}. Under native routing, $j$ is selected and $g_j^{\mathrm{cur}}>0$ for finite logits. The stated conditions then imply nonzero local expert and router-logit gradients, respectively.
\hfill$\square$

Proposition 1 shows that routing replay changes local gradient support: experts selected only by native routing lose their direct gradient contributions, whereas experts selected only by replay can receive gradients instead. Shared experts may also receive different gradients due
to changes in gate weights and downstream gradients.

\subsection{Gradient Bias Relative to Native Routing}

Let $\mathcal{R}^{\mathrm{cur}}_\theta(x,y)$ denote the routing assignments under the current training policy, and let $\mathcal{R}^{\mathrm{roll}}$ denote the recorded rollout routes. For a fixed empirical distribution $\mathcal{D}$ over training examples and their recorded routes, define
\begin{equation}
\begin{aligned}
J_{\mathrm{cur}}(\theta)
&=
\mathbb{E}_{\mathcal{D}}
\left[
\ell\left(
\theta;x,y,
\mathcal{R}^{\mathrm{cur}}_\theta(x,y)
\right)
\right],\\
J_{\mathrm{R3}}(\theta)
&=
\mathbb{E}_{\mathcal{D}}
\left[
\ell\left(
\theta;x,y,\mathcal{R}^{\mathrm{roll}}
\right)
\right].
\end{aligned}
\label{eq:r3-objectives}
\end{equation}
Both objectives use the same policy-optimization
loss, with the data, advantages, old log-probabilities,
and other loss inputs held fixed. They differ only
in the routing assignments used during the forward pass.

Away from routing selection boundaries, we define
the gradient bias relative to the native-routing
objective as
\begin{equation}
b_{\mathrm{replay}}(\theta)
=
\nabla_\theta J_{\mathrm{R3}}(\theta)
-
\nabla_\theta J_{\mathrm{cur}}(\theta).
\label{eq:r3-gradient-bias}
\end{equation}

\paragraph{Proposition 2.}
Routing replay can introduce nonzero gradient bias relative to the native-routing objective. In particular, there exist routing assignments, expert parameters, and a policy-gradient loss for which $b_{\mathrm{replay}}(\theta)\ne 0$.

This follows from Proposition~1: when native routing selects an expert excluded by replay, the two routing rules can yield different expert-parameter gradients under the same training loss.